\documentclass[letterpaper]{article} 
\usepackage[submission]{aaai24}  
\usepackage{times}  
\usepackage{helvet}  
\usepackage{courier}  
\usepackage[hyphens]{url}  
\usepackage{graphicx} 
\usepackage{natbib}  
\usepackage{caption} 
\usepackage{algorithm}
\usepackage{algorithmic}
\usepackage{amsmath}
\usepackage{xcolor}
\usepackage{multirow}
\usepackage{array}
\usepackage{newfloat}
\usepackage{listings}
\DeclareCaptionStyle{ruled}{labelfont=normalfont,labelsep=colon,strut=off} 
\floatstyle{ruled}
\newfloat{listing}{tb}{lst}{}
\floatname{listing}{Listing}
\title{Dual-Branch Temperature Scaling Calibration for Long-Tailed Recognition}
\author{
    Written by AAAI Press Staff\textsuperscript{\rm 1}\thanks{With help from the AAAI Publications Committee.}\\
    AAAI Style Contributions by Pater Patel Schneider,
    Sunil Issar,\\
    J. Scott Penberthy,
    George Ferguson,
    Hans Guesgen,
    Francisco Cruz\equalcontrib,
    Marc Pujol-Gonzalez\equalcontrib
}
\affiliations{
    \textsuperscript{\rm 1}Association for the Advancement of Artificial Intelligence\\

    1900 Embarcadero Road, Suite 101\\
    Palo Alto, California 94303-3310 USA\\
    proceedings-questions@aaai.org
}

\usepackage{bibentry}

\begin{document}

\maketitle

\begin{abstract}
The calibration for deep neural networks is currently receiving widespread attention and research. Miscalibration usually leads to overconfidence of the model. While, under the condition of long-tailed distribution of data, the problem of miscalibration is more prominent due to the different confidence levels of samples in minority and majority categories, and it will result in more serious overconfidence. To address this problem, some current research have designed diverse temperature coefficients for different categories based on temperature scaling (TS) method. However, in the case of rare samples in minority classes, the temperature coefficient is not generalizable, and there is a large difference between the temperature coefficients of the training set and the validation set. To solve this challenge, this paper proposes a dual-branch temperature scaling calibration model (Dual-TS), which considers the diversities in temperature parameters of different categories and the non-generalizability of temperature parameters for rare samples in minority classes simultaneously. Moreover, we noticed that the traditional calibration evaluation metric, Excepted Calibration Error (ECE), gives a higher weight to low-confidence samples in the minority classes, which leads to inaccurate evaluation of model calibration. Therefore, we also propose Equal Sample Bin Excepted Calibration Error (Esbin-ECE) as a new calibration evaluation metric. Through experiments, we demonstrate that our model yields state-of-the-art in both traditional ECE and Esbin-ECE metrics.
\end{abstract}

\section{Introduction}

Neural networks have achieved significant success in various fields such as image recognition\cite{krizhevsky2017imagenet,he2016deep}, object detection\cite{ren2015faster}, and semantic segmentation\cite{cordts2016cityscapes}. However, despite their impressive performance, neural network models are increasingly becoming miscalibrated. Calibration refers to the ability of a model to produce accurate and reliable predictions with good calibration confidence. In other words, a well-calibrated model should produce predictions that are consistent with the true probability of the event it predicts. Only then can the model have accurate prediction probabilities for each prediction, and we can know when the model is trustworthy. However, actual models are often miscalibrated, and this phenomenon is more pronounced under long-tail scenarios \cite{zhong2021improving}.

Under the condition of long-tail data distribution, due to the difference in the number of samples between the head and tail classes, the model actually produces more overconfidence for the head classes and less overconfidence for the tail classes \ref{fig2}. Therefore, using a uniform temperature scaling factor for all data would lead to poor calibration performance \cite{guo2017calibration}. Some studies have attempted to set different temperature coefficients for different categories, such as \cite{islam2021class}, but due to the scarcity of minority class samples, the temperature coefficients trained on the training set tend to be biased towards individual samples rather than category temperature coefficients, resulting in a large difference from the temperature coefficients that can adapt to the test set. We refer to this phenomenon as poor calibration generalization. Therefore, we propose Equal Size Bin Temperature Scaling(Esbin-TS), which divides all samples into intervals based on confidence and trains a temperature coefficient for each interval. At the same time, since we need to consider the differences between categories, there are still differences in temperature coefficients between categories, so we need to consider retaining the adaptive learning method for category temperature parameters . Finally, we design a dual-branch structure that can independently learn two temperature coefficients for the same sample based on two different settings, and fuse them by class geometric mean to obtain the final temperature coefficient for a single sample. We call this calibration framework Dual-TS.

Meanwhile, we noticed that the traditional ECE calculation method gives a higher weight to low-confidence samples in calculating the accuracy within the calculation interval, which means that low-confidence samples have a greater impact on the calibration results. To compute ECE, all
\textit{N} predictions are first grouped into \textit{B} interval bins. The traditional ECE calculation method is as follows:
\begin{equation}
    \mathrm{ECE}=\sum_{b=1}^{B} \frac{\left|\mathcal{S}_{b}\right|}{N}\left|\operatorname{acc}\left(\mathcal{S}_{b}\right)-\operatorname{conf}\left(\mathcal{S}_{b}\right)\right| \times 100 \%
\end{equation}

where $\mathcal{S}_{b}$ is the set of samples whose prediction scores fall into Bin-\textit{b},acc($\cdot$) and conf($\cdot$) are the accuracy and predicted confidence of $\mathcal{S}_{b}$, respectively.

\begin{figure}[t]
\centering
\includegraphics[width=0.9\columnwidth]{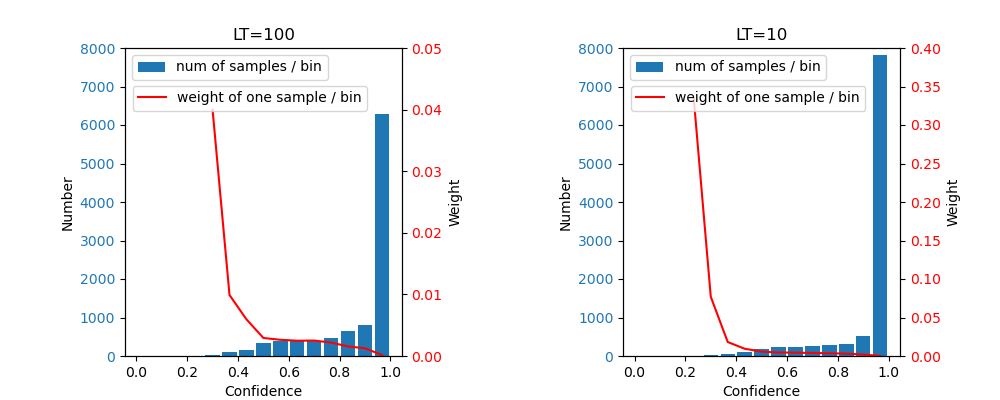}
\caption{ The total number of samples in each bin and the weight of each sample in a single bin in the CIFAR-10 dataset.}
\label{fig1}
\end{figure}

The binning method here is evenly divided based on confidence. However, since the number of samples in each confidence interval is different (Figures\ref{fig1}), the actual impact of each sample on the accuracy calculation results within each interval is different. For samples with high model confidence, which are actually reliable in model application, the assigned accuracy weight is too low. When calculating ECE with each sample having equal weight, a small number of low-confidence samples have a greater impact on the accuracy of the current bin, and therefore have a greater impact on ECE. Therefore, we designed Esbin-ECE to alleviate the excessive influence of a small number of low-confidence samples on model calibration evaluation by setting an equal number of samples in each interval.

In summary, this article has four main contributions:
\begin{itemize}
    \item We summarized and evaluated the existing category classification calibration methods under the long-tail calibration problem and proposed Class Adaptive Temperature Scaling(CA-TS).
    \item Based on the shortcomings of the category classification calibration method, we proposed a calibration method, Esbin-TS that divides all samples into equal-sized bins.
    \item We identified the shortcomings of the existing ECE evaluation method and proposed Esbin-ECE.
    \item Based on CA-TS and Esbin-TS, we designed a complete Dual-TS calibration model and demonstrated that it is the state-of-the-art in the current long-tail calibration problem.
\end{itemize}

\section{Related Work}

\subsection{Calibration for Long-tailed Recognition}

Currently, calibration can be mainly divided into two categories: in-training calibration and post-hoc calibration. In-training calibration simultaneously changes the model's accuracy and confidence. \cite{louizos2017multiplicative} proposed to introduce Bayesian neural networks to reflect the model's uncertainty and improve calibration results. \citeauthor{lakshminarayanan2017simple} proposed an ensemble method to help model calibration. \cite{zhong2021improving} further strengthened calibration by adding strategies such as mixup. \cite{mukhoti2020calibrating} improved calibration by introducing Focal Loss. \cite{kumar2018trainable} further improved the calibration effect of high-confidence samples by redefining the embedding loss with distance metrics. The first proposal of post-hoc calibration was to improve calibration without changing the model's accuracy. \cite{platt1999probabilistic} proposed a binary Plat calibration method, and \cite{guo2017calibration} extended it to multi-class calibration. Temperature scaling has since become a major method of post-hoc calibration, and subsequent articles such as \cite{ji2019bin} have improved it for better results.

Zhong et al. first proposed that when data follows a long-tail distribution, the calibration for imbalance problem becomes more severe, and suggested introducing some training strategies to alleviate the problem of calibration imbalance caused by differences in the amount of data for different categories\cite{zhong2021improving}. \cite{islam2021class} proposed an improvement to the original temperature scaling method by assigning different category calibration coefficients to each class after learning the temperature parameter with all data.  However, existing post-hoc calibration methods for long-tail data distribution have not considered the small number of samples in the tail classes and the lack of generalization of the temperature parameters for minority classes.

\subsection{Calibration Evaluation Metrics}

Currently, the evaluation of model calibration can be mainly divided into three categories: instance-level, probability-level, and non-absolute calibration metrics. Instance-level metrics include NLL Loss \cite{guo2017calibration}, Brier Score etc. These methods evaluate the model's calibration by independently calculating a result for each sample. Probability-level metrics rely on dividing all samples into intervals and then calculating the calibration metrics within each interval and aggregating the results. Common metrics include ECE, Reliability Diagram \cite{guo2017calibration}, Field-ECE \cite{pan2020field}, among which ECE is the most common data metric, while Reliability Diagram is the most common graphical metric. Non-absolute calibration metrics mainly include accuracy, recall, AUC, etc. In a typical calibration-related article, 1-2 metrics for each class are usually selected as references for comprehensive evaluation, with ECE being the most important metric. However, ECE's sample partitioning is unfair, as it focuses too much on the accuracy of low-confidence samples, which does not allow each sample to make an equal contribution to the model evaluation.

\section{Main Approach}

We denote a labeled dataset as $\mathcal{D}=\left\{\left({x}_{i}, y_{i}\right)\right\}_{i=1}^{|\mathcal{D}|}$. The training/test set are denoted by $\mathcal{D}_{tr}$, $\mathcal{D}_{te}$, respectively. Each dataset has $C$ categories. The output (logit) of each sample after the last classification layer is represented by $z_{i}=[z_{i_{1}}, z_{i_{2}}, ..., z_{i_{c}}]$, and the output after the softmax layer is represented by $p_{i}=[p_{i_{1}}, p_{i_{2}}, ..., p_{i_{c}}]$. The largest $p_{i_{c}}$ is the confidence of the sample, and the corresponding category is the predicted result, which is represented by $\hat{y_{i}}$.

\subsection{Class Adaptive Temperature Scaling}

\begin{figure}[t]
\centering
\includegraphics[width=0.9\columnwidth]{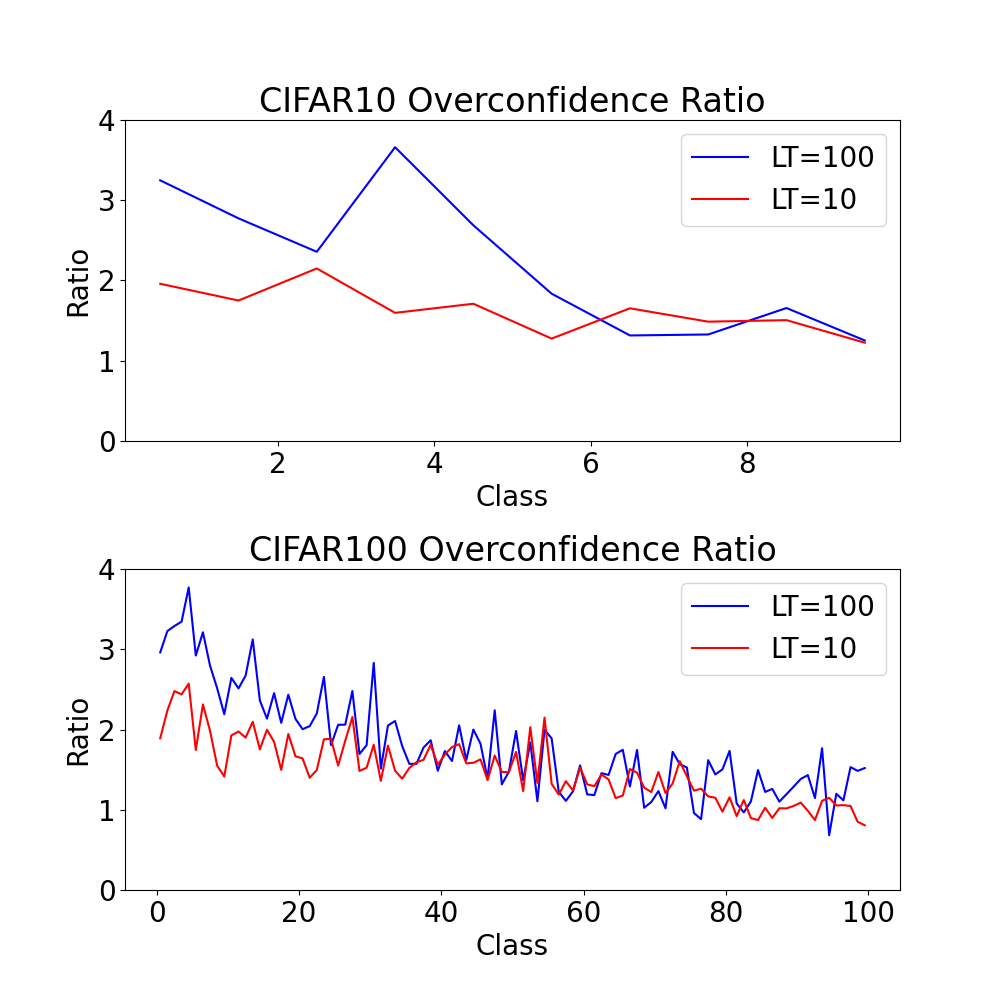}
\caption{ The theoretical temperature coefficient for each class, which is the average logit divided by how much it equals the accuracy rate ratio. Class 0 is the head class, and class numbers start from 0. As the category number increases, the sample size decreases.}
\label{fig2}
\end{figure}

Under the condition of a long-tailed distribution of data, the model tends to lean towards learning the classes with more samples, so the accuracy of the head classes is usually higher. However, the same problem is also reflected in calibration. The model gives higher confidence to the head classes, resulting in a more pronounced overconfidence effect. As shown in Figures\ref{fig2}, it can be seen that the degree of overconfidence for each head class is much higher than that of the tail classes. However, this overconfidence caused by different sample sizes cannot be simply measured by sample size alone, because there are also factors such as learning difficulty and discriminability between different classes, which also leads to the fact that the adaptive learning effect of CA-TS for each class is better than the design of a unified temperature coefficient change function, and the curve of the function can be adjusted according to different datasets in a timely manner. Therefore, we also adopt the method of adaptive learning temperature parameters for each class to provide category temperature parameters for the $z_{i}$ of each sample.

However, unlike traditional temperature calibration parameters, the temperature parameter we learn scales all the class logits of a sample. For two class logits results $z_{i_{a}}$ and $z_{i_{b}}$ of a sample, if $T_{a}$ is greater than $T_{b}$, even if $z_{i_{a}} > z_{i_{b}}$, it is still possible for $p_{i_{a}} > p_{i_{b}}$. Because $\hat{y_{i}}$ selects the prediction class $c$ by the largest $z_{i_{c}}$, this may cause changes in accuracy. However, our explanation for this is that different $T$ values represent the degree of overconfidence of the model in different classes' logits, and we provide an opportunity to re-calibrate the model's results. The formula for category-adaptive calibration is as follows:

\begin{equation}
    p_{i_{C A}}=\sigma_{S M}\left(\frac{\left[z_{i_{1}}, z_{i_{2}}, \ldots, z_{i_{c}}\right]}{\left[T_{1}^{C A}, T_{2}^{C A}, \ldots, T_{c}^{C A}\right]}\right)
\end{equation}

where $\sigma{SM}$ denotes Softmax layer, $p_{i_{CA}}$ denotes the confidence of sample i calibrated by Class Adaptive Temperature Scaling, $\left[T_{1}^{C A}, T_{2}^{C A}, \ldots, T_{c}^{C A}\right]$ denotes the temperature of $C$ classes respectively.

\subsection{Equal Size Bin Temperature Scaling}

We sort all samples according to their confidence $p_{i} = \sigma_{S M}(z_{i})$, and divide all samples into B sub-datasets with equal sample sizes for each bin, denoted as $D_{B}\left(\left|D_{i}\right|=\left|D_{j}\right|\right)$ . The samples within a bin form a sub-dataset, and we ensure that each sub-dataset has the same sample size while keeping the confidence of the samples as similar as possible. Similar to traditional temperature calibration, we scale all the logits of each sample based on the bin it belongs to. Since the logits between classes of the same sample are scaled using the same temperature coefficient, Esbin-TS does not affect the judgment of the prediction results, nor does it affect the accuracy of the model.

Due to the long-tailed distribution of the data, the learning of the tail class parameter $T_{t}^{CA}$ in CA-TS does not have generalization. We hope to build larger sub-datasets with similar sample characteristics for the tail class samples to assist in training temperature coefficients. Based on this idea, we constructed Esbin-TS.

Equal Size Bin Temperature Scaling can be expressed by the following formula:

\begin{equation}
    p_{i_{E S}}=\sigma_{S M}\left(\frac{z_{i}}{T_{b}^{E S}}\right), p_{i} \in D_{b}
\end{equation}

where $D_{b}$ denotes that the \textit{i}-th sample belongs to the \textit{b}-th sub-dataset, $T_{b}^{E S}$ denotes the temperature of sub-dataset $D_{b}$, $p_{i_{E S}}$ denotes the confidence calibrated by Equal Size Bin Temperature Scaling.

\subsection{Dual-Branch Temperature Scaling}

\begin{figure*}[t]
\centering
\includegraphics[width=1.8\columnwidth]{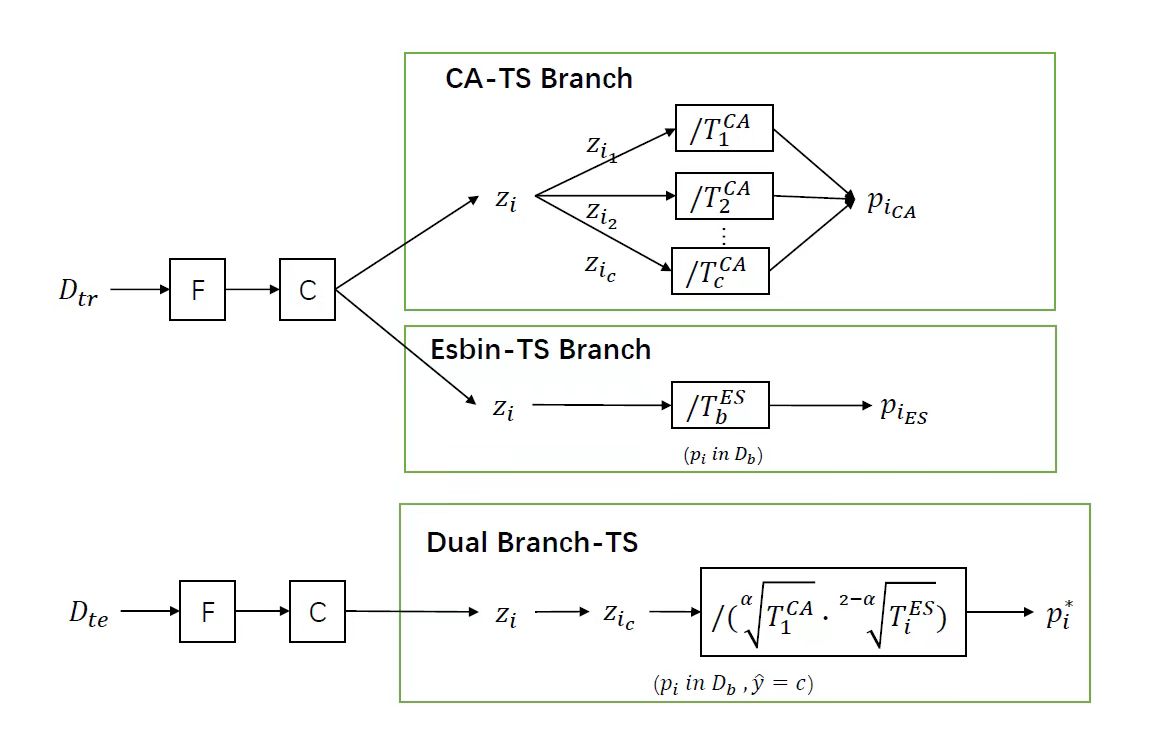}
\caption{ $F$ represents the feature extraction layer, and $C$ represents the final classification layer. The training dataset($D_{tr}$)samples are respectively processed by the CA-TS Branch and the Esbin-TS branch to train $T_{c}^{CA}$ and $T_{b}^{ES}$. The fused $T$ is used for calculation in the test dataset($D_{te}$).}
\label{fig3}
\end{figure*}

In the case of long-tailed data, through the discussions in sections Esbin-TS and CA-TS, we have learned that there are advantages to both adapting the temperature coefficient based on class and setting the temperature coefficient for equal sample bins based on similar confidence principles. However, traditional temperature calibration methods have always chosen one of these methods to scale the samples. Here, we hope to design an architecture that can integrate multiple calibration frameworks, so that multiple calibration methods can complement each other's shortcomings and play to their respective strengths. We believe that the simplest and most effective way is to average all the temperature coefficients obtained from each branch for each sample, and use the averaged temperature parameter as the final calibration result. In this paper, based on CA-TS and Esbin-TS, we designed a dual-branch temperature calibration model, as shown in Figure \ref{fig3}.

In the training set, we train two branches and use the trained temperature parameters to obtain calibrated results in the test set. Specifically, in the CA-TS branch, we set a class temperature calibration coefficient for each class, and train each class of $z_{i}$ passed through the neural network model, finally obtaining the temperature calibration coefficient $\left[T_{1}^{C A}, T_{2}^{C A}, \ldots, T_{c}^{C A}\right]$ corresponding to each class. In the Esbin-TS branch, we select the training temperature parameter for each sample based on the confidence subset $D_{b}$ to which the sample belongs, and finally obtain the Esbin temperature coefficient $\left[T_{1}^{E S}, T_{2}^{E S}, \ldots, T_{b}^{E S}\right]$. In the test dataset, we sort all samples by confidence and select the Esbin temperature coefficient used for each sample, while judging the predicted label $\hat{y}$ of the sample and selecting the class temperature calibration coefficient used. For \textit{i}-th sample, its final confidence is as follows:

\begin{equation}
    p_{i}^{*}=\sigma_{S M}\left(\frac{z_{i}}{\sqrt[\alpha]{T_{c}^{C A}} \cdot \sqrt[2-\alpha]{T_{b}^{E S}}}\right), \hat{y}=c \text { and } p_{i} \in D_{b}
\end{equation}

where $\alpha$ is a hyperparameter, and we will discuss its selection and role in detail in Sec Experiment.

\section{Experiments}

\subsection{Datasets}

We selected cifar-10 and cifar-100 \cite{krizhevsky2009learning}as our datasets, and designed two types of imbalance factors for each dataset, as shown in \cite{cao2019learning}.

\subsection{Inplemention Details}

For the cifar dataset, we chose ResNet18 \cite{zhang2017mixup}as the main framework for the neural network model. We used SGD gradient descent with a batch size of 1024 and a momentum of 0.9, and trained for 200 epochs in the training phase. For the entire posterior calibration phase, we used NLL loss and LBFGS optimizer to optimize the temperature parameters. For data distributions with LT of 10, we used an initial temperature coefficient of 1.5, and for data with LT of 100, we used an initial temperature coefficient of 2.0.

\begin{table*}[htbp]
    \centering
    \caption{Top-1 accuracy ($\%$) / ECE ($\%$) / Esbin-ECE ($\%$) and NLL Loss for ResNet-18 based models trained on CIFAR-10-LT and CIFAR-100-LT. \textbf{Bold} representss the best result and \underline{underline} represents the second best result (Esbin-ECE expected)}.
    \label{table1}
    \resizebox{2.1\columnwidth}{!}{
    \begin{tabular}{|>{\rule[-1ex]{0pt}{3.5ex}}c|c c c c|c c c c|}
        \hline
        \multirow{3}{*}{Method} & \multicolumn{4}{|c|}{CIFAR-10-LT} & \multicolumn{4}{|c|}{CIFAR-100-LT} \\
        \cline{2-9}
        \multirow{3}{*}{} & \multicolumn{2}{|c|}{IF10} & \multicolumn{2}{|c|}{IF100} & \multicolumn{2}{|c|}{IF10} & \multicolumn{2}{|c|}{IF100} \\
        \cline{2-9}
        \multirow{3}{*}{} & ACC / ECE / Esbin-ECE&NLL& ACC / ECE / Esbin-ECE&NLL& ACC / ECE / Esbin-ECE&NLL& ACC / ECE / Esbin-ECE&NLL\\
        \hline
        baseline & 84.14 / 9.23 / 9.21 & 0.677 & 63.84 / 24.8 / 24.8 & 1.748 & 57.61 / 18.0 / 18.0 & 1.925 & 38.57 / 32.8 / 32.8 & 3.183 \\
        \cline{2-9}
        TS & 84.14 / 3.78 / - & \underline{0.522} & 63.84 / 7.46 / -& 1.071 & 57.61 / 4.29 / - & \underline{1.698} & 38.57 / 3.66 / - & \underline{2.541} \\
        \cline{2-9}
        CDA-TS & \underline{84.37} / \underline{3.58} / - & \underline{0.522} & \underline{64.20} / \underline{6.66} / -& \underline{1.061} & \textbf{57.93} / \underline{3.67} / - & \textbf{1.682} & \underline{38.99} / \underline{3.10} / - & \textbf{2.527} \\
        \cline{2-9}
        Dual-TS & \textbf{85.17} / \textbf{1.20} / \textbf{1.16} & \textbf{0.512} & \textbf{70.29} / \textbf{2.68} / \textbf{2.67} & \textbf{1.059} & \underline{57.71} / \textbf{1.81} / \textbf{1.81} & 1.700 & \textbf{42.33} / \textbf{2.61} / \textbf{2.51} & 2.547 \\
        \cline{2-9}
        \hline    
    \end{tabular}
    }
\end{table*}

\begin{figure*}[t]
\centering
\includegraphics[width=2.0\columnwidth]{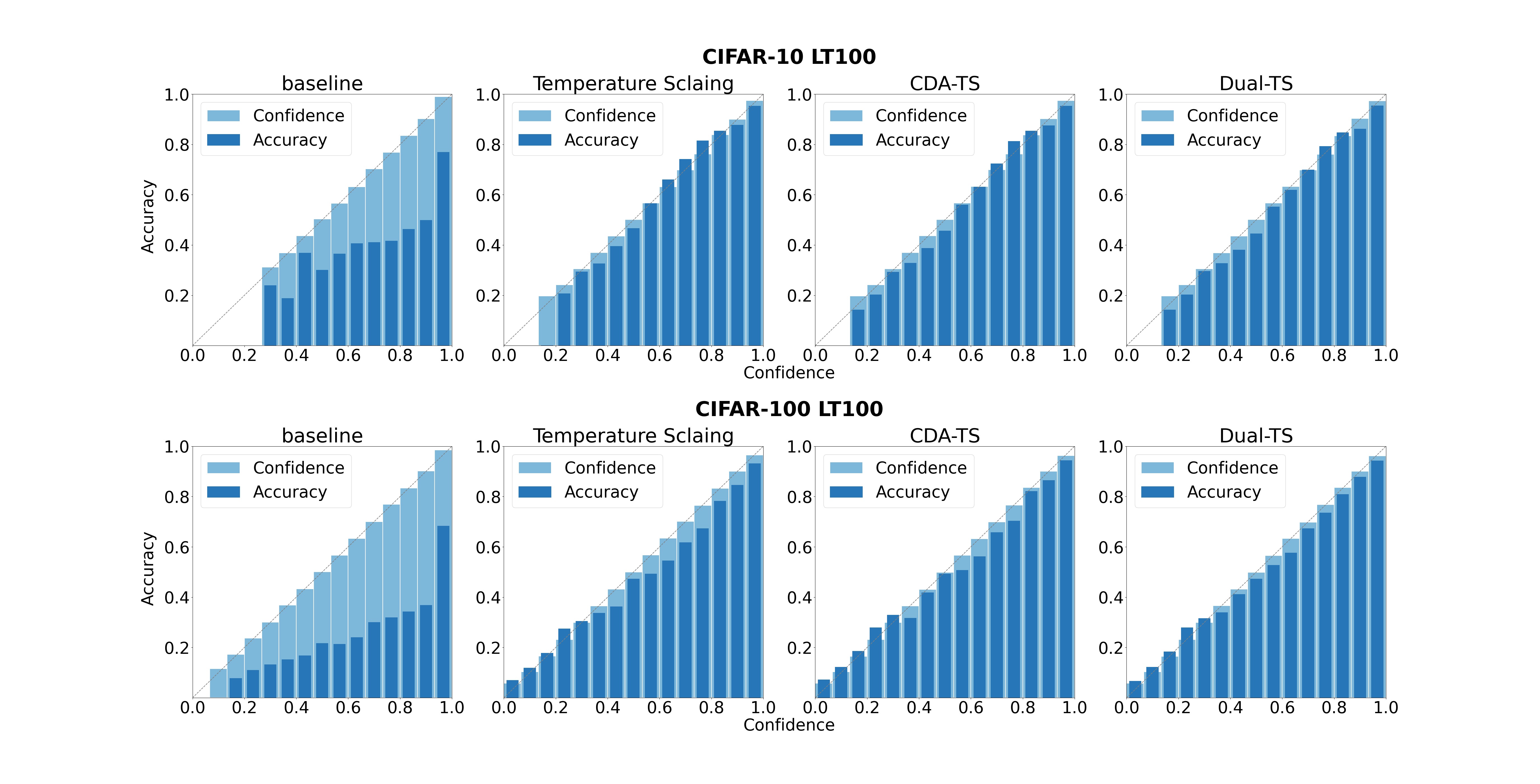}
\caption{ Reliability diagrams of ResNet-18 trained on CIFAR-10-LT with IF 100 and CIFAR-100-LT with IF 100. From left to right: baseline, Temperature Scaling, CDA-TS and Dual-TS. The datasets selected from top to bottom are CIFAR-10 LT100 and CIFAR-100 LT100.}
\label{fig4}
\end{figure*}

\subsection{Evaluation Metrics}

As mentioned in section related work, we chose one of three evaluation metrics (accuracy, NLL loss, ECE) as the model's evaluation standard. However, we noticed that the most common ECE calibration metric is not entirely accurate in evaluating the model. Although the calculation result of ECE gives equal weight to each sample, it gives too much weight to low-confidence samples when calculating the accuracy of a single bin, as shown in Figure\ref{fig1}. Therefore, we designed Esbin-ECE as another calibration evaluation metric. Its calculation method is shown below:

\begin{equation}
    \operatorname{acc}\left(D_{b}\right)=\frac{1}{\left|D_{b}\right|} \sum_{i \in D_{b}} 1\left(\widehat{y}_{i}=y_{i}\right)
\end{equation}
\begin{equation}
    \operatorname{conf}\left(D_{b}\right)=\frac{1}{\left|D_{b}\right|} \sum_{i \in D_{b}} {p}_{i}
\end{equation}
\begin{equation}
    \operatorname{Esbin-ECE}=\sum_{b=1}^{B} \frac{\left|D_{b}\right|}{N}\left|\operatorname{acc}\left(D_{b}\right)-\operatorname{conf}\left(D_{b}\right)\right| \times 100 \%
\end{equation}

where $\left|D_{1}\right|=\left|D_{2}\right|=\cdots=\left|D_{b}\right|$ and $\forall p_{1} \in D_{1}, p_{2} \in D_{2}, \ldots, p_{b} \in D_{b} ; \exists p_{1} \leq p_{2} \leq \cdots \leq p_{b}$.

Esbin-ECE divides all samples into $B$ intervals based on confidence, with an equal number of samples in each interval. This ensures that each sample contributes equally to the average accuracy and average confidence in their respective intervals, avoiding situations where accuracy and confidence in low-confidence intervals are represented by only a few samples.

\begin{table}[htbp]
    \centering
    \caption{ECE ($\%$) and Esbin-ECE ($\%$) for ResNet-18 based models trained on CIFAR-10-LT and CIFAR-100-LT.}
    \label{table2}
    \resizebox{1.0\columnwidth}{!}{
    \begin{tabular}{|c|c c c c|c c c c|}
        \hline
        \multirow{3}{*}{Method} & \multicolumn{4}{|c|}{CIFAR-10-LT} & \multicolumn{4}{|c|}{CIFAR-100-LT} \\
        \cline{2-9}
        \multirow{3}{*}{} & \multicolumn{2}{|c|}{IF10} & \multicolumn{2}{|c|}{IF100} & \multicolumn{2}{|c|}{IF10} & \multicolumn{2}{|c|}{IF100} \\
        \cline{2-9}
        \multirow{3}{*}{} & ECE & Esbin-ECE & ECE & Esbin-ECE & ECE & Esbin-ECE & ECE & Esbin-ECE \\
        \hline
        baseline & 9.23 & 9.21 & 24.8 & 24.8 & 18.0 & 18.0 & 32.8 & 32.8 \\
        \cline{2-9}
        CA-TS & 2.59 & 2.40 & 2.11 & 2.16 & 3.78 & 4.17 & 4.01 & 4.02 \\
        \cline{2-9}
        Esbin-TS & 1.48 & 1.54 & 2.70 & 2.65 & 1.90 & 2.17 & 3.49 & 3.64 \\
        \cline{2-9}
        Dual-TS & 1.20 & 1.16 & 2.67 & 2.64 & 1.81 & 2.12 & 2.61 & 2.51 \\
        \cline{2-9}
        \hline    
    \end{tabular}
    }
\end{table}

\subsection{Results}

As shown in table\ref{table1}, our dual-branch calibration model achieved the best performance on each metric for each dataset.  Our Dual-TS model achieved the best performance on the ECE metric, while showing significant improvements in accuracy compared to all previous models, except for the SOTA CDA on the CIFAR-100 lt10 dataset. The NLL metric is relatively high on the CIFAR-100 dataset, but we have achieved good calibration performance, indicating that Du al-TS still has the potential to further improve its calibration performance. Figure\ref{fig4} shows the comparison results of the reliability diagrams. From the figure, we can see that we have achieved better calibration performance for the low-confidence samples, while effectively avoiding over-calibration for the high-confidence samples. We will provide more data on Esbin-ECE in the next section.

\subsection{Ablation Experiments}

\subsubsection{Performance of Class Adaptive Calibration and Equal Size Bin Calibration}

\begin{figure}[t]
\centering
\includegraphics[width=0.9\columnwidth]{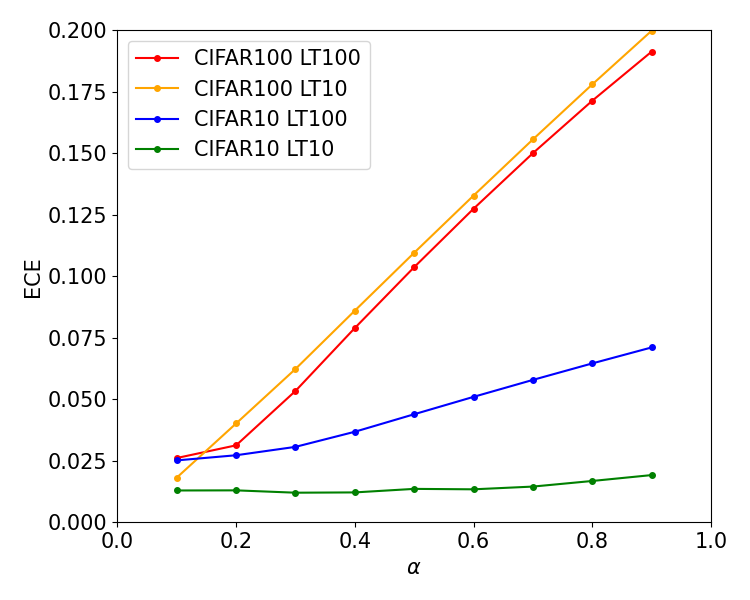}
\caption{ ECE for different $\alpha$ trained on CIFAR-10-LT and CIFAR-100-LT.}
\label{fig5}
\end{figure}

To test the independent effects of the two branches, we conducted ablation experiments on both branches. The experimental results are shown in Table \ref{table2}, where we found that both branches had good independent effects, but the combined effect of the two branches was better. This indicates that both branches can provide good calibration solutions for the vast majority of samples, but each branch sacrifices the best calibration effect for a portion of the samples. Dual-TS Calibration combines the advantages of both branches to achieve better results.

\subsubsection{Choose for $\alpha$}

In the fusion of the dual branches, we need to choose the value of $\alpha$. We have discussed that both branches are effective, and the value of $\alpha$ actually represents which branch the model should be more biased towards, that is, the effectiveness of each branch. We chose an interval of 0.1 for $\alpha$, and the experimental results are shown in Figure \ref{fig5}.

\section{Conclusion}

In this paper, we discussed the calibration problem in the context of long-tailed data distribution. We proposed Esbin-TS as a novel temperature scaling method, analyzed the Esbin-TS and CAD-TS branches, and designed the complete Dual-TS framework. We also identified the shortcomings of the traditional ECE calculation method and proposed Esbin-ECE as a new calibration metric. Finally, we demonstrated that our Dual-TS framework achieved the state-of-the-art performance on existing long-tailed data calibration problems.

\bigskip

\bibliography{aaai24}

\end{document}